\def\year{2021}\relax
\documentclass[letterpaper]{article} 
\usepackage{aaai21}  
\usepackage{times}  
\usepackage{helvet} 
\usepackage{courier}  
\usepackage[hyphens]{url}  
\usepackage{graphicx} 
\usepackage{graphicx}  
\usepackage{natbib}  
\usepackage{caption} 
\usepackage{amsmath, amssymb}
\usepackage{subfig}
\usepackage{algorithm}
\usepackage{algorithmic}
\usepackage{makecell, multirow, tabularx, booktabs}
\newcommand{\B}{\mathcal{B}}
\newcommand{\E}{\mathbb{E}}
\DeclareMathOperator*{\argmax}{arg\,max}
\DeclareMathOperator*{\argmin}{arg\,min}
\title{ADER:Adapting between Exploration and Robustness for Actor-Critic Methods}
\author{Bo Zhou, Kejiao Li, Hongsheng Zeng, Fan Wang, Hao Tian}
\affiliations{
    Baidu., Inc
}

\begin{document}

\maketitle
\vspace{-10cm}
\begin{abstract}

Combining off-policy reinforcement learning methods with function approximators such as neural networks has been found to lead to overestimation of the value function and sub-optimal solutions. Improvement such as TD3 has been proposed to address this issue. However, we surprisingly find that its performance lags behind the vanilla actor-critic methods (such as DDPG) in some primitive environments. 
In this paper, we show that the failure of some cases can be attributed to insufficient exploration. We reveal the culprit of insufficient exploration in TD3, and propose a novel algorithm toward this problem that ADapts between Exploration and Robustness, namely ADER. To enhance the exploration ability while eliminating the overestimation bias, we introduce a dynamic penalty term in value estimation calculated from estimated uncertainty, which takes into account different compositions of the uncertainty in different learning stages. Experiments in several challenging environments demonstrate the supremacy of the proposed method in continuous control tasks.
\end{abstract}

\section{Introduction}
Recent studies have revealed that combining off-policy reinforcement learning (RL) with function approximators (such as neural networks) can lead to systematic overestimation bias of value function \cite{hasselt2010double,van2016deep}. To address this problem, several methods have been proposed to reduce the overestimation through introducing the supplementary value function for target value computation such as Double Q Learning (DDQN) \cite{hasselt2010double}, and TD3 \cite{fujimoto2018addressing}. The latter one is one of the most widely used actor-critic methods for its simplicity and effectiveness in continuous control problems, which takes the minimum estimated value given by two value functions as the target value. In spite of its success in many cases, we found that it lagged behind the vanilla actor-critic methods such as DDPG \cite{lillicrap2015continuous}, in both asymptotic performance and sample-efficiency, in some primitive environments. 

\begin{figure*}
    \centering
        \subfloat[]{\includegraphics[width=0.20\textwidth]{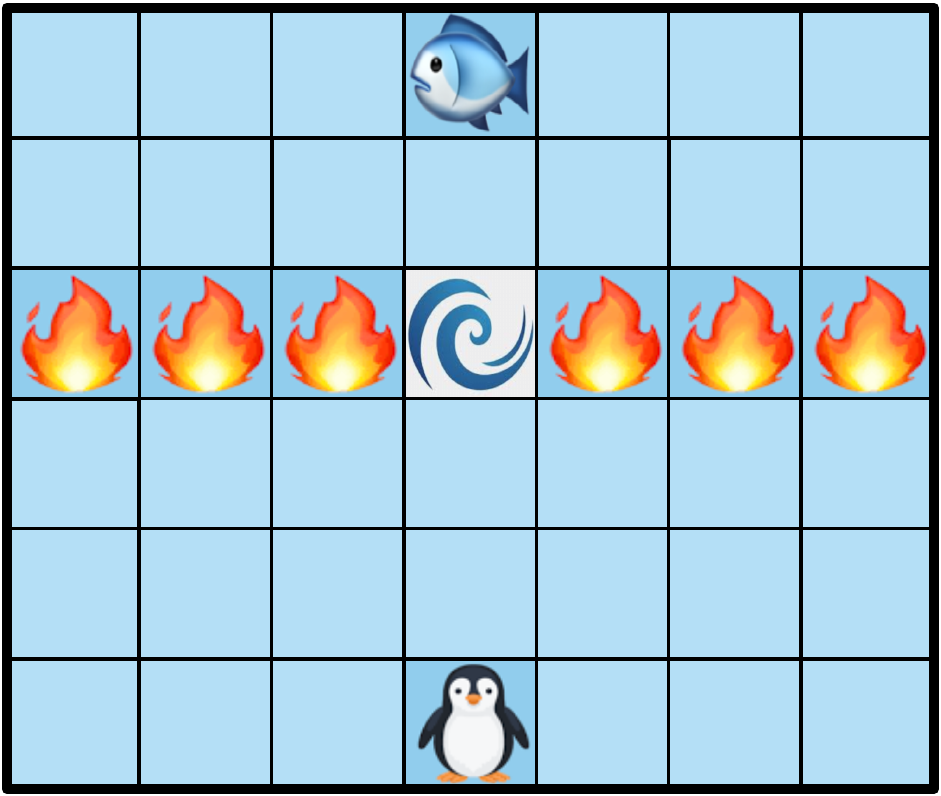}} 
        \subfloat[]{\includegraphics[width=0.46\textwidth]{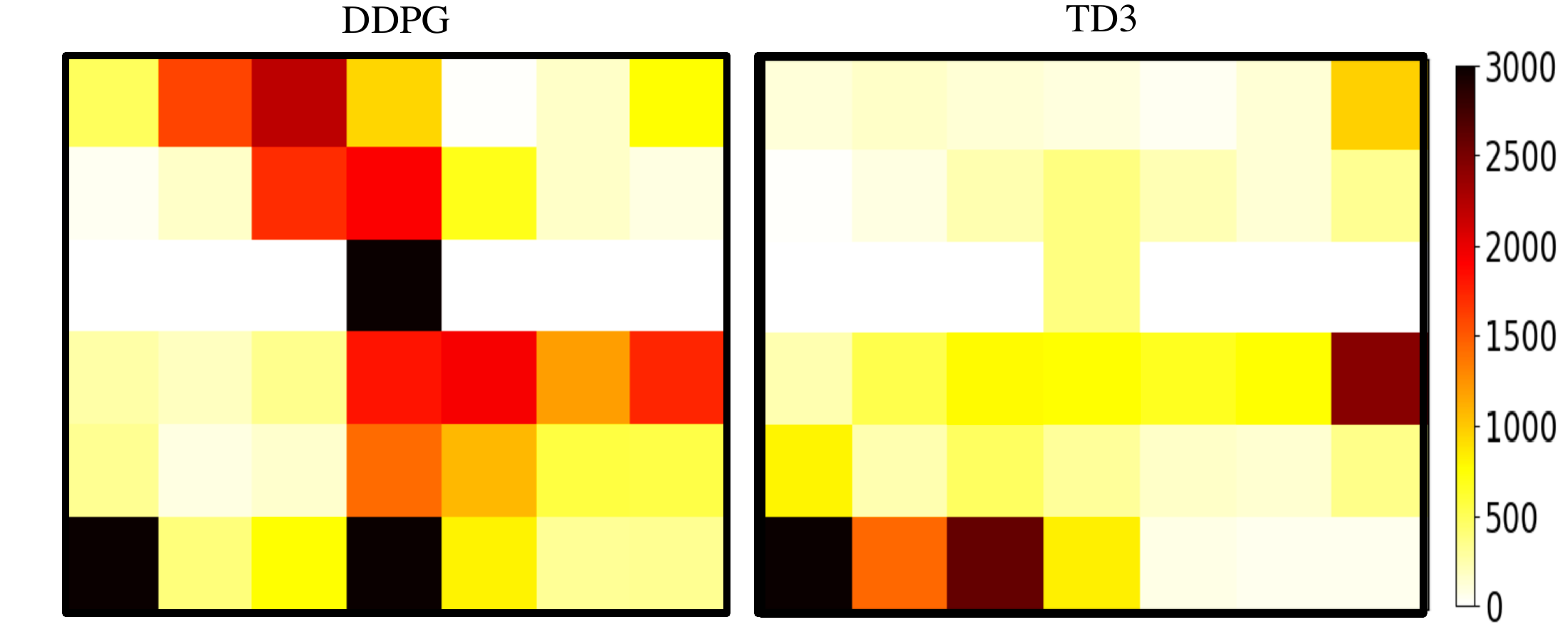}} 
        \subfloat[]{\includegraphics[width=0.25\textwidth]{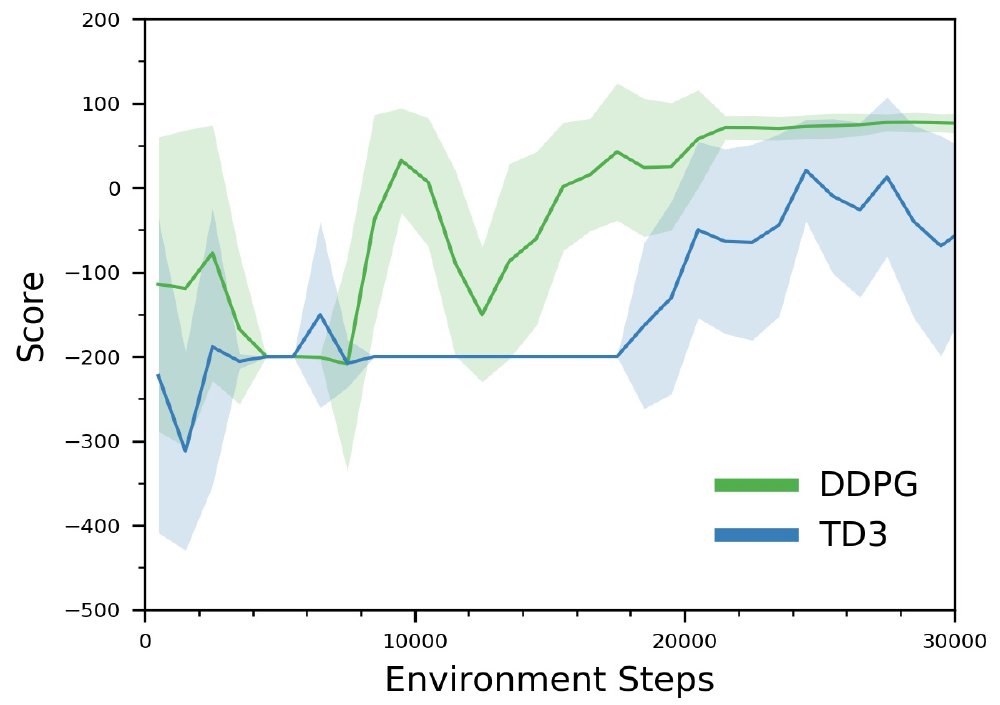}} 
        \caption{A counter example showing the exploration issue of TD3. (a) A grid world environment where a penguin tries to catch the fish at the top. It is blocked by a fire wall, and the only path is through the storm grid, which may randomly push it onto adjacent grids. (b) The visitation count of each grid of DDPG and TD3 during the training process. (c) The performance of DDPG and TD3 on the grid world environment.}
\label{Fig:ToyCase}
        
\end{figure*}

In Figure~\ref{Fig:ToyCase}(a) we show a toy case with a penguin at the bottom of the grid world trying to catch the fish at the top. A wall of fire blocks it, and the only way is through the storm grid in the center. The transitions in all the grids are clean and deterministic, except the storm grid that leads the penguin to a random neighboring grid that is irrelevant to its decision (For complete environment settings, please refer to Appendix~\ref{Sec:Details for Toy Env}). On this grid environment, we evaluated the TD3 and DDPG algorithms. The experimental result in Figure~\ref{Fig:ToyCase}(c) shows that TD3 performance falls behind DDPG. By inspecting the behaviors of the agents, we found that the agent trained with TD3 has a lower probability of reaching the global optimum compared with DDPG, which is to cross the storm to reach the target. We plot the visiting count of each grid during the whole training process in Figure~\ref{Fig:ToyCase}(
b). The figure shows that TD3 has a substantially lower visiting count to the storm grid than DDPG, which prevents the agent from reaching the goal.

Concluding from the above case, TD3 discourages some key explorations to certain extent, which has prevented it from achieving higher asymptotic performance. We argue that the \textit{minimum} operation in TD3 conflicts with the principle of \textit{optimism in the face of uncertainty}, which is related to uncertainty of estimation \cite{lai1985asymptotically, strehl2005theoretical}. This means that TD3 reduces the overestimation bias of the value function at the cost of insufficient exploration, which can lead to sub-optimal policies. We will present the full analysis in Section~\ref{Sec:exploration issue}.

In this paper, we propose a novel algorithm called ADER, considering the estimation uncertainty in value estimation. Estimation uncertainty has mainly two components \cite{dabney2018implicit,mavrin2019distributional}: \textit{parametric uncertainty}, which arises from prediction variance of function approximators given finite training samples, and \textit{intrinsic uncertainty}, which arises from the inherent stochasticity of the environment. The parametric uncertainty decreases as more training samples are acquired, while the intrinsic uncertainty persists along with the training. Both kinds of uncertainty may lead to estimation bias, but parametric uncertainty acts a more important role for exploration. Although it is difficult to distinguish the different components precisely, we can introduce an automatic tuning schedule to emphasize different uncertainties at different learning stages: at the early learning stage when data is insufficient, minor penalty on parametric uncertainty can facilitate exploration, and at the later training stage when the intrinsic uncertainty dominates estimation uncertainty, high penalty on intrinsic uncertainty can reduce the overestimation bias. In this way, we are able to retain efficient exploration and reduce overestimation bias at the same time. We evaluated the algorithm in several challenging continuous control environments. Experimental results show that our algorithm benefits both exploration efficiency and policy robustness, achieving better performance than the existing RL methods.

\section{Related Work}
Exploration with optimism in the face of uncertainty has been widely adopted in deep RL methods. The count-based approach \cite{tang2017exploration} estimates the uncertainty with the state visitation counts, which are then used as the reward bonus. Curiosity-driven exploration \cite{pathak2017curiosity,badia2020agent57} introduces a reward bonus for exploration that is related to how much the agent knows the environment.  Bootstrapped DQN \cite{osband2016deep} employs a multi-headed neural network to represent the ensemble Q-values to perform exploration. The variance of the estimated values given by different heads can also be used as a metric of the estimation uncertainty. Among these methods, the estimation uncertainty is employed as a reward bonus, but our investigation on TD3 shows that its update is taking the estimation uncertainty as a penalty for reward shaping.

There are two sources of uncertainty associated with exploration in model estimation.: \textit{parametric uncertainty} and \textit{intrinsic uncertainty} \cite{lai1985asymptotically,dabney2018implicit,mavrin2019distributional}. While the parametric uncertainty decreases as the size of training data increases, the intrinsic uncertainty persists along with the training as it arises from the inherent systematic stochasticity of the environment. There are some methods that \cite{moerland2017efficient} seek to combine both types of uncertainty to improve exploration, with Bayesian
drop-out \cite{gal2016dropout} for parametric uncertainty estimation, and Gaussian distribution for intrinsic uncertainty estimation. While it is hard to quantify different types of uncertainty, the distributional RL method \cite{mavrin2019distributional} suppresses the intrinsic uncertainty by introducing a decaying schedule for robust policy learning, avoiding selecting the actions with higher variance.

There have also been other works of exploration improvement over actor-critic methods. Soft Actor-Critic (SAC) \cite{haarnoja2018soft} performs policy learning with entropy maximization, which can acquire diverse behaviors and improve exploration. Optimistic Actor-Critic (OAC) \cite{ciosek2019better} collects the training data with a behavior policy derived from the training policy for optimistic exploration and performs the off-policy update. Following the principle of optimism in the face of uncertainty. The behavior policy was obtained by maximizing the approximate of the upper confidence bound of the state-action value estimation. The main difference between OAC and our work is that we improve the exploration efficiency of the training policy, while OAC optimizes the behavior policy.

\section{Insufficient Exploration Issue in TD3}
\label{Sec:exploration issue}

\subsection{Actor-Critic Methods}
In Reinforcement learning(RL), an agent learns to interact with the environment \cite{sutton2018reinforcement} following Markov Decision Processes (MDP). We use $s \in S$ to denote the state, and $a \in A$ to denote the action, which depends on the policy $\pi(a | s)$. The environment returns a reward $r$, and informs the next state $s'$ following the transition $p(s' | a, s)$, which is normally unknown to the agent. The goal of the problem is to find the optimal policy $\pi^*$ that maximizes the expected future reward: $\pi^* = \max\limits_{\pi} \mathbb{E}_{a\sim\pi,s'\sim p} \left[ \sum_{t=0}^{\infty} \gamma^t r_t\right]$, where $\gamma \in (0, 1) $ is the discount factor.

The state-action value function under policy $\pi$ is defined as 
$Q(s, a) = \E_{a\sim\pi,s'\sim p} \left[ \sum_{t=0}^{\infty} \gamma^t r_t \right | s_0=s, a_0=a]$. In Q-learning \cite{watkins1992q}, the state-action value function $Q$ is updated by minimizing the temporal difference (TD) error:
\begin{align}
\mathcal{L}_{\text{TD}} =  (y - Q(s,a))^2
\label{EQ:temporal_difference_loss}
\end{align}
\begin{align}
y = r + \max\limits_{a' \in A}\gamma Q(s',a')
\label{EQ:q_learning_target}
\end{align}
DQN \cite{mnih2015human} further uses the neural network as the function approximator in Q-learning algorithm. The authors proposed two improvements to enhance the convergence stability: experience replay buffer and target network. The transitions $(s, a, r, s')$ are stored in a first-in-first-out (FIFO) queue named experience replay buffer, and samples will be sampled uniformly and repeatedly for training. The target network $Q_{\theta^-}$ maintains a delayed copied parameters of the training network $Q_{\theta}$. The target values are computed with the target network:

\begin{equation}
y_{\text{DQN}} = r + \gamma \max\limits_{a' \in A} Q_{\theta^-}(s',a')
\label{EQ:TD3}
\end{equation}

For continuous control problems, it is necessary to introduce a policy network $\pi_{\phi}(a|s)$, which formed the actor-critic method. For instance, the deep deterministic policy gradient (DDPG) \cite{lillicrap2015continuous} updates the policy network by maximizing the state-action value function as follow,
\begin{equation}
\phi = \phi - \alpha \nabla_{\pi_\phi} Q(s, a),
\end{equation}
where the state-action pair $(s,a)$ is uniformly sampled from the experience replay buffer. For stable training with neural networks, DDPG also maintains a pair of target networks for actor and critic networks ($Q_{\theta^-}, \pi_{\phi^-}$) that slowly track the training parameters: $\theta^- \leftarrow \tau \theta + (1 - \tau) \theta^-, \phi^- \leftarrow \tau \phi + (1 - \tau) \phi^-$, with $\tau \ll 1$.
\subsection{Overestimation Bias}
The maximum operator of Equation~\ref{EQ:q_learning_target} lead to systematic overestimation bias, due to the unavoidable prediction error $\epsilon$ occurs in estimated values, considering the following fact \cite{thrun1993issues}:
\begin{equation}
\E_{\epsilon}\left[\max\limits_{a'\in A} \{Q_\theta(s',a') + \epsilon\}\right] \geqslant\E\left[\max\limits_{a'\in A} Q_\theta(s', a')\right].
\label{EQ:Overestimation_Bias}
\end{equation}
To address this problem, Double DQN \cite{van2016deep} uses two value functions to disentangle action selection from value estimation, one for value estimation, and another for action selection. The overestimation bias persists in actor-critic methods such as DDPG \cite{fujimoto2018addressing}. The TD3 algorithm employs two distinct neural networks ($Q_{\theta_1}, Q_{\theta_2}$) for value estimation, taking the minimum between the two for calculating the target value:
\begin{equation}
y_{\text{TD3}} = r + \gamma \min\limits_{i=1,2}Q_{\theta_i^-}(s',\pi_{\phi^-}(s'))
\label{EQ:TD3}
\end{equation}

\subsection{The Issue of Insufficient Exploration}
With a few calculations, the target value in Equation~\ref{EQ:TD3} can be represented in the following form: 
\begin{align}\label{EQ:TD3_core}
y_{\text{TD3}} &= r + \gamma min(Q_1, Q_2) \nonumber\\
  &= r - \gamma \sigma(Q_1, Q_2) + \gamma \mu(Q_1, Q_2),
\end{align}
where $\mu$ and $\sigma$ represent the statistical mean and standard variance respectively. We provide the derivation in Appendix~\ref{Sec:TD3 derivation}. Equation~\ref{EQ:TD3_core} shows that the \textit{minimum} operation in TD3 is equal to approximating the lower confidence bound of the value estimation by subtracting the uncertainty from the mean value.

On the other hand, effective exploration has been one of the main challenges in deep reinforcement learning. Recent work on uncertainty-based exploration shows that exploration guided by \emph{estimation uncertainty} can enhance the exploration efficiency \cite{bellemare2016unifying}, which employs the following target:
\begin{equation}
y_{\text{EXP}} = r + \beta N(s,a)^{-\frac{1}{2}} + \gamma \max\limits_{a' \in A} (Q(s', a')),
\label{EQ:exploration}
\end{equation}
where $N(s, a)$ denotes the visiting count of the state action pair $(s,a)$. 
Previous theory has revealed that the parametric uncertainty of any estimation at some point $s'$ is proportional to the reciprocal of the square root of the visiting count \cite{auer2002finite,koenker2005econometric}, and considering the visiting count of $(s,a)$ is proportional to that of the state $s'$ by $p(s'|s,a)$, the second term in the right-hand side of Equation~\ref{EQ:exploration} is also proportional to the parametric uncertainty of the value function at $s'$. This is also called the principle of \emph{optimism in the face of uncertainty}.

As the stand derivation of the estimates given by ensemble models are often used as the metric of estimating uncertainty \cite{smith2001disentangling,osband2016deep,lakshminarayanan2017simple}, comparing Equation~\ref{EQ:TD3_core} and Equation~\ref{EQ:exploration} it is not hard to find out why TD3 lead to the insufficient exploration problem. While TD3 uses estimation uncertainty as a punishment in the target value to address the overestimation bias, it discourages exploring the states with high parametric uncertainty. Thus, a key contribution of our work is to solve this conflict to achieve better performance.

\section{Methodology}
Generally, estimation uncertainty is composed of the parametric and intrinsic uncertainty, and it is difficult to distinguish them from each other. However, these two sources of uncertainty have different trends in training. We thus introduce an automatic schedule to dynamically adjust the penalty on estimation uncertainty, which can benefit the exploration at the early stage of training and reduce the overestimation bias at the later stage. We describe the algorithm for ADER in Section 4.1, and justify it in Section 4.2.

\subsection{Formulation}
For convenience we are considering the value update for transition $s,a,s'$. We use $\mu$, $\sigma$ to denote the mean and uncertainty of $(Q_1, Q_2)$ respectively. We further use $\sigma_I$ and $\sigma_P$ to denote the intrinsic and parametric uncertainty respectively. Note that we have $\sigma = \sigma_I + \sigma_P$. Following the aforementioned analysis, the ideal computation of target values can be written as:
\begin{align}
    y_{\text{ideal}} &= r + \gamma( \mu - \alpha * \sigma_I + \beta * \sigma_P),
\label{EQ:Ideal}
\end{align} 
where $\alpha>0$ determines the penalty on intrinsic uncertainty to reduce the overestimation bias, and $\beta>0$ indicates the bonus on parametric uncertainty for exploration. This allows the algorithm to explore the state space following the parametric uncertainty and to reduce the overestimation bias introduced by the intrinsic uncertainty at the same time. 
Unfortunately, we can not precisely estimate $\sigma_P$ and $\sigma_I$ separately. However, it is possible to approximate $\sigma$ through an ensemble of neural networks. Formally, we define the ADER algorithm as using the following target value:
\begin{align}
    y_{\text{ADER}} &= r + \gamma (\mu - \eta(t) * \sigma), \nonumber \\
    \eta(t) &=  \alpha - \kappa\sqrt{\frac{\log t}{t}}.
\label{EQ:ADER_schema}
\end{align}
where $\alpha$ and $\kappa$ are constant values, $t$ is an epoch counter starting from $t=2$ and updated every K environment steps. By setting $\alpha = 1$ and $\kappa = 0$, ADER degenerates to TD3. With $\alpha = 0$ and $\kappa = 0$, ADER degenerates to the DDPG algorithm with two ensemble Q-value functions.

\subsection{Interpreting ADER}
Starting from the target value computation in TD3, for an arbitrary state-action pair, we wish to find the $\eta(t)$ to adjust the penalty on estimation uncertainty such that the target values computed by Equation~\ref{EQ:ADER_schema} is identical to the values computed by Equation~\ref{EQ:Ideal}:
\begin{align}
    y_{\text{ADER}} = y_{\text{ideal}}
\label{EQ:ADER EQ IDEAL}
\end{align}
Then the resulting $\eta(t)$ has the following approximate form: 
\begin{align}
\eta(t) = \alpha - (\alpha + \beta)\frac{\sigma_P}{\sigma_I}
\label{EQ:proportional}
\end{align}
The proof is deferred to Appendix~\ref{Sec:ADER derivation}. Since $\sigma_P$ rises from the environment property and remains constant, the right term of Equation~\ref{EQ:proportional} is proportional to $\sigma_P$: $(\alpha + \beta)\frac{\sigma_P}{\sigma_I} \propto \sigma_P$. Given that the decay of parametric uncertainty is statistically proportional to the rate: $\sigma_P \propto  c\sqrt{\frac{logt}{t}}$ \cite{koenker2005econometric,mavrin2019distributional}, where c is a constant value, we approximately replace the right term in Equation~\ref{EQ:proportional} with $c\sqrt{\frac{logt}{t}}$, and we can get: $\eta(t) \approx  \alpha - \kappa \sqrt{\frac{logt}{t}}$, where $\kappa=(\alpha + \beta)c$.

Thus ADER introduces two additional hyper-parameters: $\alpha$ and $\kappa$. Note that $\alpha$ determines the penalty on intrinsic uncertainty in the later training stage: $\lim\limits_{t \to \infty} \eta(t) = \alpha$. A higher value of $\alpha$ can benefit the policy robustness, since the penalty on intrinsic uncertainty encourages the deployment policy to be less likely to select the actions with high variance in future return. Once we set the value of $\alpha$, $\kappa$ indicates how fast the bonus of parametric uncertainty for exploration decays.

\begin{figure}[b] 
\centering
\subfloat[]{\includegraphics[width=0.25\linewidth]{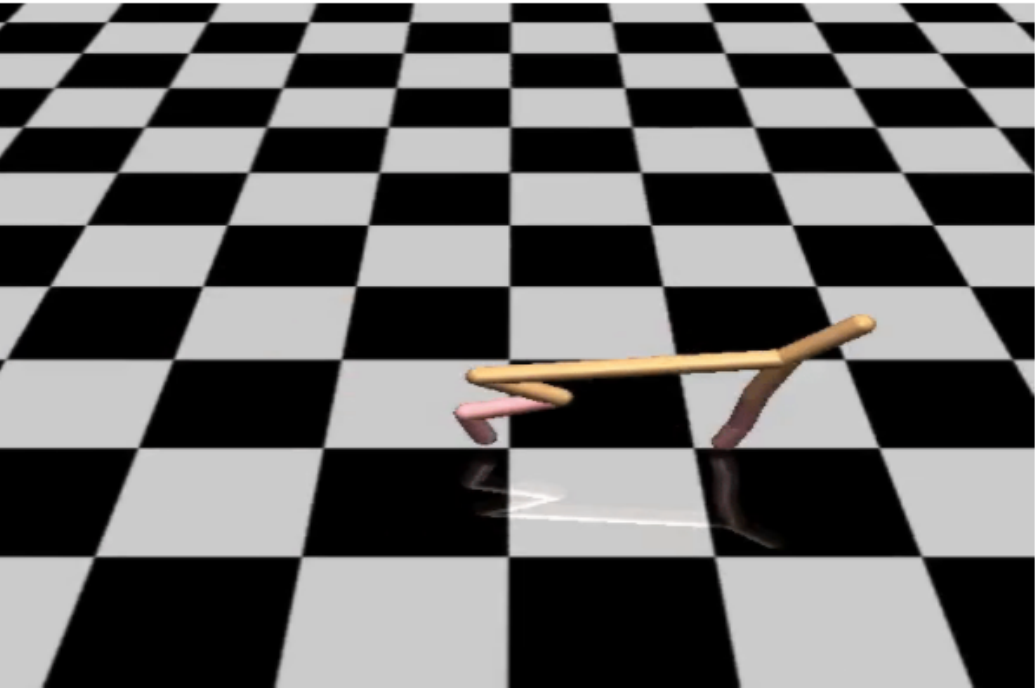}}
\subfloat[]{\includegraphics[width=0.25\linewidth]{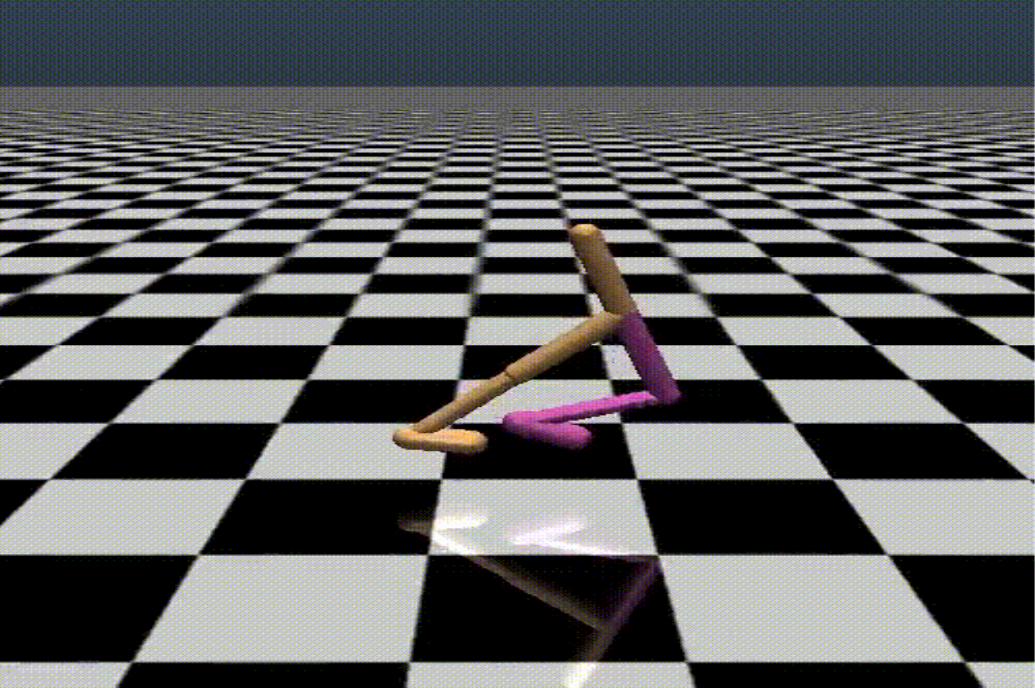}}
\subfloat[]{\includegraphics[width=0.25\linewidth]{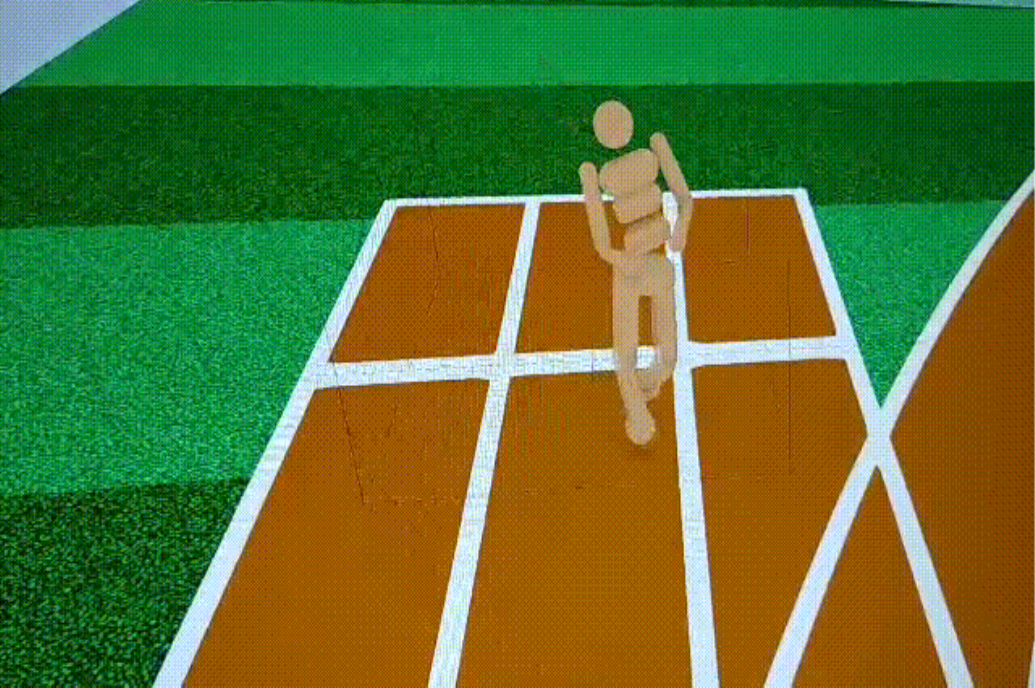}}
\subfloat[]{\includegraphics[width=0.25\linewidth]{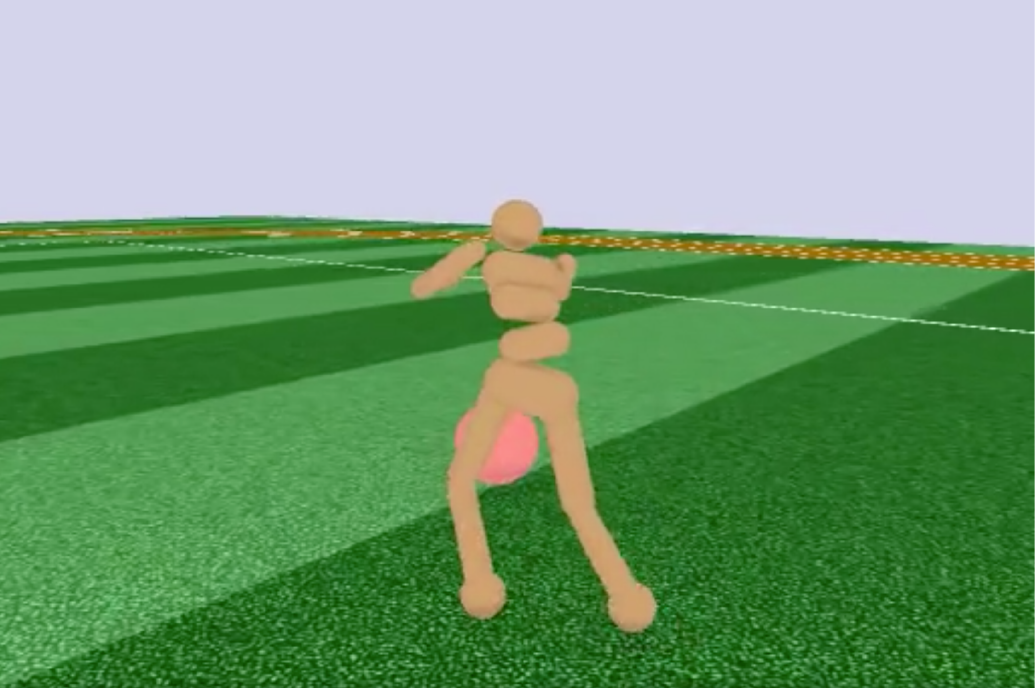}}
\caption{Example experiment environments (a) HalfCheetah-v1, (b) Hopper-v1, (c) RoboschoolHumanoid-v1, (d) RoboschoolHumanoidFlagrun-v1.}
\label{fig:env}
\end{figure}


\subsection{Training Stability}
Reinforcement learning methods have been known to be unstable when combined with nonlinear function approximators (such as neural networks), and it is essential to maintain the target networks to stabilize the training \cite{mnih2015human, lillicrap2015continuous}. As the value of $\alpha$ depends on the environment steps, it changes rapidly and makes the distribution of target values unstable. We follow the update of target networks in DQN \cite{mnih2015human} and update the $\alpha$ periodically. That is, we update the value of $\alpha$ every $M$ train steps, rather than update it instantly after each epoch. We provide a full algorithm description in Algorithm~\ref{alg:ADER}.

    \begin{algorithm}
      \caption{ADER} \label{alg:ADER}
      \begin{algorithmic}[1]
        \REQUIRE Initialize actor network : $\pi_\phi$ , and\\
        2 critic networks: \{$Q_{\theta 1}, Q_{\theta 2}$\} \\
        Initialize target networks: $\phi^- \leftarrow \phi, \theta_1^- \leftarrow  \theta_1, \theta_2^- \leftarrow  \theta_2$ \\
        Initialize experience replay buffer $\B$ \\
        Initialize epoch counter t=2
        \WHILE{not done}
        \STATE Collect transitions $\{s, a, r, s', a'\}$ using the policy $\pi_\phi$, and append the data into $\B$ \\
        \STATE update the epoch counter every K steps:  $t \leftarrow t+1$ \\
        \STATE update $\eta(t)$ periodically according to the Equation~\ref{EQ:ADER_schema} periodically.
        \FOR{$\theta_i$ \;\text{in} \; $\{Q_{\theta 1}, Q_{\theta 2}\}$}
            \STATE sample a mini-batch of transitions from $\B$
            \STATE compute target values y:
            \STATE \hspace{0.5cm} $\{y_1, y_2\} = r + \gamma Q_{\theta i}^-(s', \pi_{\theta^-}(s'))$
            \STATE \hspace{0.5cm}  $\sigma_y = \sigma(\{y_1, y_2\})$
            \STATE \hspace{0.5cm}  $\overline{y} = \mu(\{y_1, y_2\})$
            \STATE \hspace{0.5cm}  $y = \overline{y} - \eta(t) * \sigma_y$
            \STATE update critic: $\theta_i \leftarrow \argmin_{\theta_i} (Q(s, a) - y)^2$
        \ENDFOR
        \STATE sample a mini-batch of transitions from $\B$
        \STATE update actor: $\phi \leftarrow \argmax_\phi (Q_{\theta_1}(s, \pi_\phi(s))$
        \STATE update target networks:
        \STATE \hspace{0.5cm} $\theta^-_i \leftarrow \tau \theta_i + (1 - \tau) \theta^-_i$
        \STATE \hspace{0.5cm} $\phi^- \leftarrow \tau \phi + (1 - \tau) \phi^-$
        \ENDWHILE
      \end{algorithmic}
    \end{algorithm}

\begin{figure*}[t]
\centering
\includegraphics[width=\linewidth]{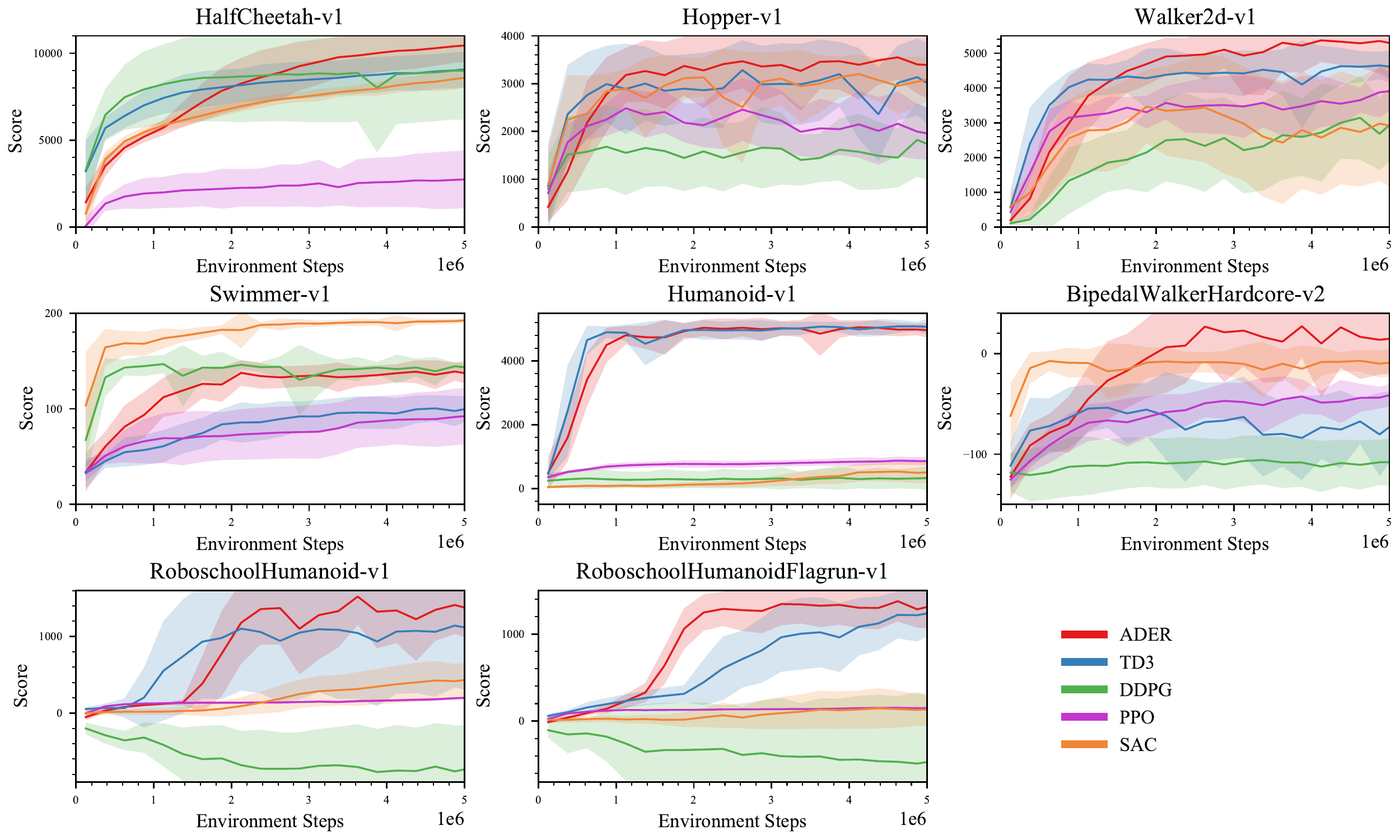}\\
\caption{Learning curves for the OpenAI continuous control tasks. The shaded region represents half a standard deviation of the average evaluation over 10 trials. Curves are smoothed uniformly for visual clarity.} 
\label{fig:OpenAIcurves}
\vspace{-.5cm}
\end{figure*}

\section{Experiments}
We evaluate ADER on Mujoco environments \cite{todorov2012mujoco}, and Roboschool \cite{Klimov2017} environments (Figure~\ref{fig:env}). The baseline algorithms are state-of-the-art RL algorithms in the continuous control domain, including PPO \cite{schulman2017proximal}, TD3, and SAC \cite{haarnoja2018soft}. First, we compare the proposed algorithm against baselines in 8 challenging tasks. Then we perform ablations on each one of the improvements we propose in ADER. We also provide an additional analysis showing that ADER can enhance policy robustness and exploration. Each experiment was run four times with different random seeds.

\subsection{Implementation and Results}
\label{sec:experiments}
Our implementation of ADER is based on the TD3 implementation provided by the authors. We keep all the hyper-parameters the same as TD3, except for the additional hyper-parameters introduced by ADER. The selection of additional hyper-parameters is obtained by the grid search method \cite{larochelle2007empirical}, with search range of $\alpha$ being $\left[0, 3\right]$ with step 0.3, and the search range of $\kappa$ being $\left[1, 10\right]$ with step 1.0.
For an exhaustive description of the hyper-parameters, please refer to Appendix~\ref{sec:hyper-parameters}. For implementations of the baseline algorithms, we used the code provided by the authors. Note that for fair comparison in sample efficiency, following the experiment setting in TD3, we train exactly 1 iteration step for all the off-policy algorithms, while the authors of SAC reported the experimental results with 4 iteration step. To study how the exploration efficiency affects the asymptotic performance, we run the experiment in each environment with 5 million steps, far more than the 1 million steps used in TD3 experiments.

We present the experimental results in Figure~\ref{fig:OpenAIcurves}. ADER achieves the best average performance except for Swimmer-v1 and Humanoid-v1. For most environments, ADER falls behind TD3 in the first 1 million steps, but it soon surpasses the performance of TD3. It shows that ADER has paid some cost at the early training stage to encourage more exploration, but this could indeed help to achieve better performance in the subsequent stage. It is also worth noticing that both TD3 and ADER are successful in quite challenging environments such as Humanoid-v1, RoboschoolHumanoid-v1, and RoboschoolHumanoidFlagrun-v1, where the  control dimension is more challange, and the agent is prone to fall over. In contrast,  without addressing the approximation error, DDPG has the poorest performance in those environments. Thus the main performance gain over DDPG can probably be attributed to the improvement of policy robustness by addressing the approximation error. Also, on-policy PPO has shown poor sample efficiency and thus lags behind the other off-policy algorithms in many cases.

\begin{table*}[ht]
\centering
\begin{center}
\begin{small}
\begin{tabularx}{\textwidth}{|c|X|X|X|X|X|}
\toprule
\bf{Environment} & \bf{Basic} & \bf{TD3} & \bf{No-RI} & \bf{No-PU} & \bf{HPS} \\
\midrule
HalfCheetah-v1 	& \bf{10476}& 9559 		& 9916	& 10416 		& 10868  \\
Hopper-v1 			& \bf{3476 } & 2585 	& 2433		& 3282		& 3579 \\
Walker2d-v1 		& \bf{5346 } & 4170		& 4796 		& 5269	& 5393 \\
Swimmer-v1 		& \bf{120 } & 100		& \bf{120} 		& 118	& 128 \\
Humanoid-v1 			& 5000 & 4965 		& \bf{5324}		& 4997	& 5324 \\
BipedalWalkerHardcore-v2 		& -72	& -92	   	& \bf{104} 	& -68		& 104\\
RoboschoolHumanoid-v1 	& 1441 	& 1091 	& 470  & \bf{1457}	& 1481  \\
RobschoolHumanoidFlagrun-v1 	& \bf{1311} 	& 1232 	& 584	& 1257	& 1311 \\
\bottomrule
\end{tabularx}
\end{small}
\end{center}
\vspace{-.2cm}
\caption{Average return of ADER variants on tested environments. The maximum value for each task is bolded, except the \textbf{HPS} variant. Most of the performance gain comes from dynamically adjusting the penalty on estimation uncertainty, and the performance can be further improved by searching a proper combination of hyper-parameters of $\alpha$ and $\kappa$.}
\label{Tab:albation}
\vskip -0.1in
\end{table*}

\subsection{Ablation Study}
To understand how each specific component affects the performance, we perform ablative experiments and compare the following variants of ADER:
\begin{itemize}
\item \textbf{TD3}: We remove the penalty adaption schedule and set $\alpha=1, \kappa=0$ in ADER, which degenerates to TD3.
\item \textbf{Basic}: $\alpha=2, \kappa=5$, a basic setting for ADER.
\item \textbf{No-RI}: $\alpha=1, \kappa=5$, using a smaller $\alpha$ as TD3 to remove \textbf{R}obustness \textbf{I}mprovement.
\item \textbf{No-PU}: The basic setting of ADER without \textbf{P}eriodical \textbf{U}pdate of $\eta(t)$.
\item \textbf{HPS} or \textbf{H}yper-\textbf{P}arameter \textbf{S}earch. ADER with best combination of $\alpha$ and $\kappa$ using the grid search method.
\end{itemize}

We present the ablation results in Table~\ref{Tab:albation}. The largest performance gain comes from the step from TD3 to Basic, which demonstrates the importance of dynamic adjustments through $\eta(t)$. Periodically updating the $\alpha$ is also essential for stabilizing the training process. The No-PU group falls behind the basic group in nearly all the tested environments except in RoboshoolHumanoid-v1, which is consistent with the ablative experiments on the target network reported in DQN \cite{mnih2015human}. After setting a smaller value for $\alpha$ (No-RI), we observe the performance drop in five environments such as HalfCheetah-v1 and Hopper-v1. We argue that a small $\alpha$ will lead the policy to favor with risky actions and collapse the performance, while a large $\alpha$ discourages the exploration. We will further discuss on the selection of $\alpha$ in the following part. The last column obtaining superior performance shows that the performance can be further improved by searching the combination of two hyper-parameters of $\alpha$ and $\kappa$ specifically for each environment. One possible reason is that the decay rate of parametric uncertainty varies from environment to environment, due to the different environmental complexity.

\begin{figure} 
\centering
\vspace{-0.2cm}
\includegraphics[width=1.0\linewidth]{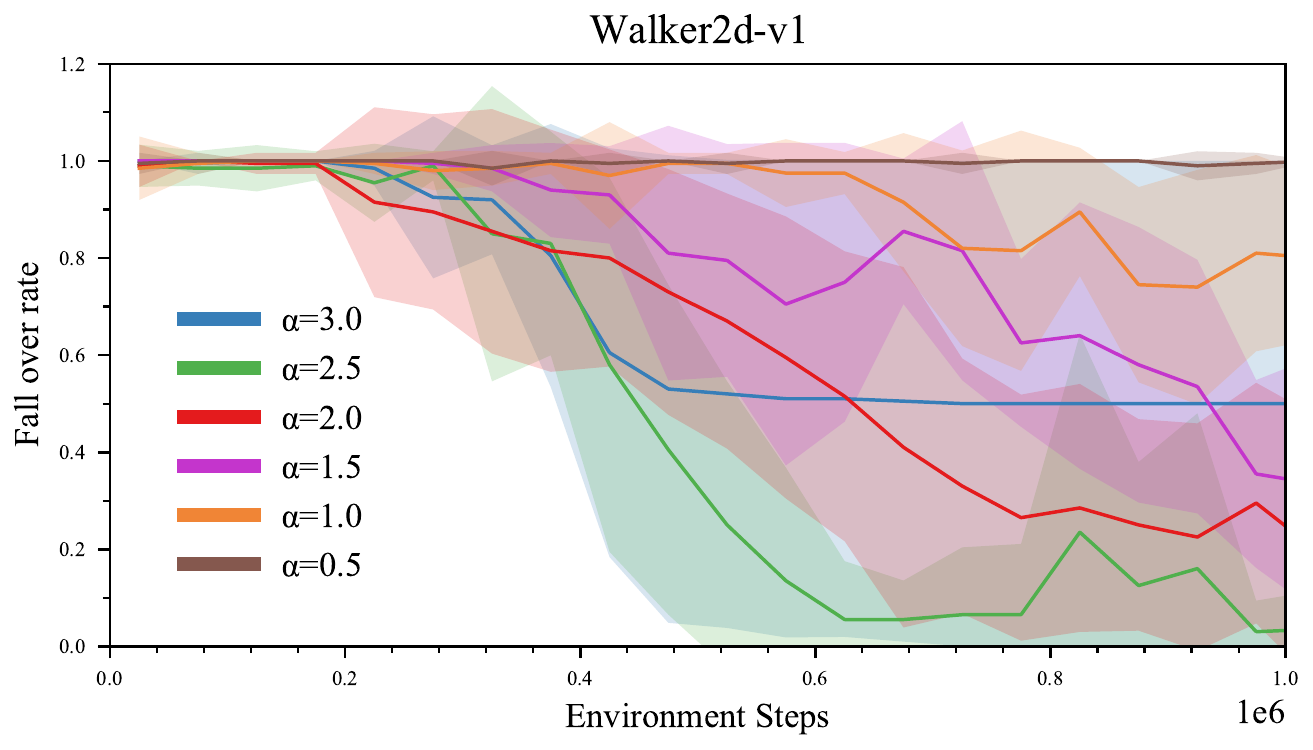}
\vspace{-.3cm}
\caption{The policy learned with higher $\alpha$ is more robust than those learned with lower $\alpha$. Each learned policy was evaluated with 10 episodes at each point. The fall over rate indicates how many times the agent runs steadily without falling over in 10 episodes.}
\label{Fig:robustness}
\vspace{-.5cm}
\end{figure}

To further investigate how increasing $\alpha$ helps the robustness, we studied more variants of ADER with $\alpha$ given different values in the Walker2d-v1 environment. Rather than use environmental score as a metric, we plot the \emph{fall over rate} against the environment steps, resulting in Figure~\ref{Fig:robustness}. It turns out that the fall over rate decreases steadily as we increase $\alpha$ before $\alpha < 2.5$, but further increasing $\alpha$ to 3.0 becomes counter-productive as exploration is severely suppressed.

\subsection{Exploration Efficiency}
\begin{figure}[t]
\centering
\includegraphics[width=1.0\linewidth]{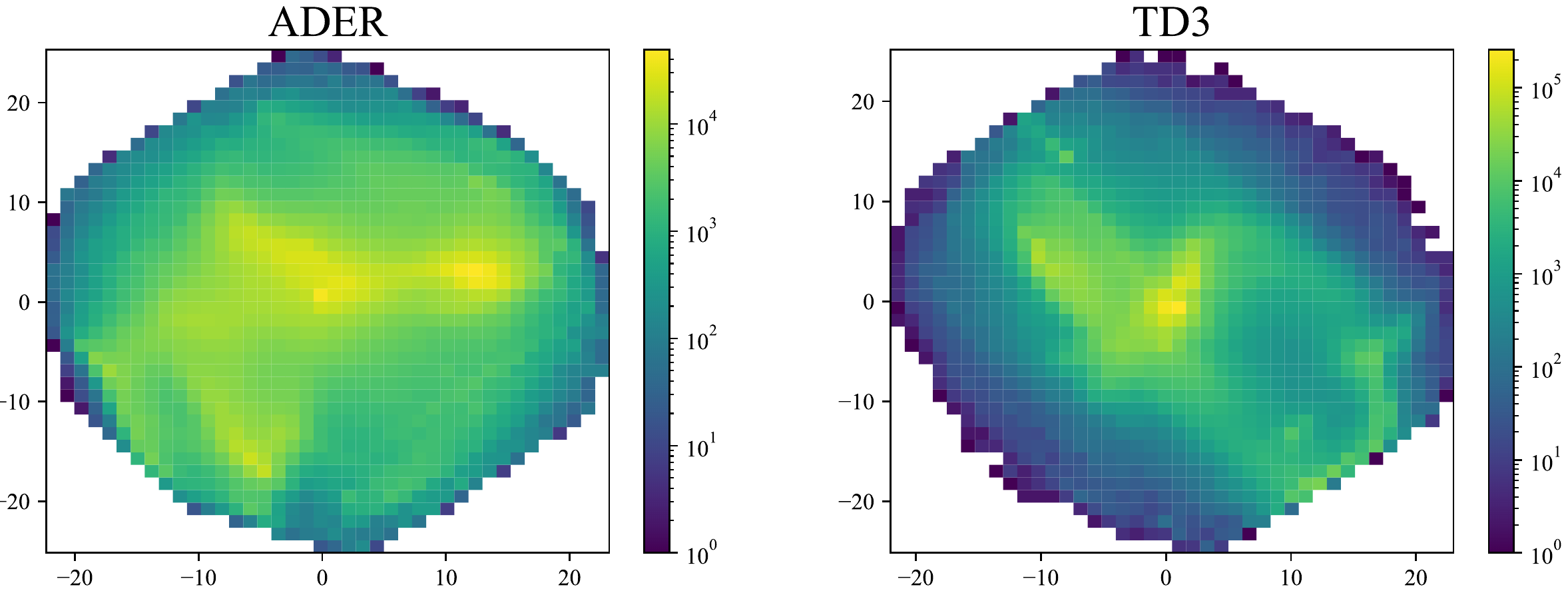}
\caption{The visitation counts of the states collected in the training of the Walker2d-v1 environment. We project all the states into a 2d plane to visualize the state distribution.}
\label{Fig:exploration_heatmap}
\vspace{-0.4cm}
\end{figure}

To validate that ADER facilitates the exploration, we study the diversity of visited states during the training process. We collect the 5 million states during the experiments reported in Section~\ref{sec:experiments} and projected them into a 2d plane to visualize the state distribution (Figure~\ref{Fig:exploration_heatmap}), using the principal component analysis (PCA) to reduce the state dimension \cite{jolliffe2016principal}. In the figure of ADER, the bright points are distributed evenly in the plane, while most of the points of TD3 are concentrated in the center. It indicates that ADER can lead to a richer set of states and improve exploration.

\section{Conclusion}
In this work, we research the insufficient exploration problem of TD3, which can lead to locally optimal solutions. Inspired by the principle of optimism in the face of uncertainty, we present the ADER algorithm as an improvement. ADER uses an automatic schedule to deal with two sources of estimation uncertainty, which encourages exploration at the early training stage and reduces the overestimation bias at the later stage. Empirically, we show that ADER achieves better performance than state-of-the-art RL methods while benefiting the exploration and policy robustness. A possible extension of this work could be state-dependent tuning of the penalty factors. Also, as uncertainty estimation is important for both exploration and policy robustness, it is of interest to precisely model the parametric and intrinsic uncertainty separately. \\


\clearpage

\bibliography{aaai}

\begin{thebibliography}{30}
\providecommand{\natexlab}[1]{#1}
\providecommand{\url}[1]{\texttt{#1}}
\providecommand{\urlprefix}{URL }
\expandafter\ifx\csname urlstyle\endcsname\relax
  \providecommand{\doi}[1]{doi:\discretionary{}{}{}#1}\else
  \providecommand{\doi}{doi:\discretionary{}{}{}\begingroup
  \urlstyle{rm}\Url}\fi

\bibitem[{Auer, Cesa-Bianchi, and Fischer(2002)}]{auer2002finite}
Auer, P.; Cesa-Bianchi, N.; and Fischer, P. 2002.
\newblock Finite-time analysis of the multiarmed bandit problem.
\newblock \emph{Machine learning} 47(2-3): 235--256.

\bibitem[{Badia et~al.(2020)Badia, Piot, Kapturowski, Sprechmann, Vitvitskyi,
  Guo, and Blundell}]{badia2020agent57}
Badia, A.~P.; Piot, B.; Kapturowski, S.; Sprechmann, P.; Vitvitskyi, A.; Guo,
  D.; and Blundell, C. 2020.
\newblock Agent57: Outperforming the atari human benchmark.
\newblock \emph{arXiv preprint arXiv:2003.13350} .

\bibitem[{Bellemare et~al.(2016)Bellemare, Srinivasan, Ostrovski, Schaul,
  Saxton, and Munos}]{bellemare2016unifying}
Bellemare, M.; Srinivasan, S.; Ostrovski, G.; Schaul, T.; Saxton, D.; and
  Munos, R. 2016.
\newblock Unifying count-based exploration and intrinsic motivation.
\newblock In \emph{Advances in neural information processing systems},
  1471--1479.

\bibitem[{Ciosek et~al.(2019)Ciosek, Vuong, Loftin, and
  Hofmann}]{ciosek2019better}
Ciosek, K.; Vuong, Q.; Loftin, R.; and Hofmann, K. 2019.
\newblock Better exploration with optimistic actor critic.
\newblock In \emph{Advances in Neural Information Processing Systems},
  1787--1798.

\bibitem[{Dabney et~al.(2018)Dabney, Ostrovski, Silver, and
  Munos}]{dabney2018implicit}
Dabney, W.; Ostrovski, G.; Silver, D.; and Munos, R. 2018.
\newblock Implicit quantile networks for distributional reinforcement learning.
\newblock \emph{arXiv preprint arXiv:1806.06923} .

\bibitem[{Fujimoto, Van~Hoof, and Meger(2018)}]{fujimoto2018addressing}
Fujimoto, S.; Van~Hoof, H.; and Meger, D. 2018.
\newblock Addressing function approximation error in actor-critic methods.
\newblock \emph{arXiv preprint arXiv:1802.09477} .

\bibitem[{Gal and Ghahramani(2016)}]{gal2016dropout}
Gal, Y.; and Ghahramani, Z. 2016.
\newblock Dropout as a bayesian approximation: Representing model uncertainty
  in deep learning.
\newblock In \emph{international conference on machine learning}, 1050--1059.

\bibitem[{Haarnoja et~al.(2018)Haarnoja, Zhou, Abbeel, and
  Levine}]{haarnoja2018soft}
Haarnoja, T.; Zhou, A.; Abbeel, P.; and Levine, S. 2018.
\newblock Soft actor-critic: Off-policy maximum entropy deep reinforcement
  learning with a stochastic actor.
\newblock \emph{arXiv preprint arXiv:1801.01290} .

\bibitem[{Hasselt(2010)}]{hasselt2010double}
Hasselt, H.~V. 2010.
\newblock Double Q-learning.
\newblock In \emph{Advances in neural information processing systems},
  2613--2621.

\bibitem[{Jolliffe and Cadima(2016)}]{jolliffe2016principal}
Jolliffe, I.~T.; and Cadima, J. 2016.
\newblock Principal component analysis: a review and recent developments.
\newblock \emph{Philosophical Transactions of the Royal Society A:
  Mathematical, Physical and Engineering Sciences} 374(2065): 20150202.

\bibitem[{Klimov and Schulman(2017)}]{Klimov2017}
Klimov, O.; and Schulman, J. 2017.
\newblock Roboschool.
\newblock \url{https://github.com/openai/roboschool}.

\bibitem[{Koenker(2005)}]{koenker2005econometric}
Koenker, R. 2005.
\newblock Econometric Society Monographs: Quantile Regression.
\newblock \emph{New York: Cambridge University} .

\bibitem[{Lai and Robbins(1985)}]{lai1985asymptotically}
Lai, T.~L.; and Robbins, H. 1985.
\newblock Asymptotically efficient adaptive allocation rules.
\newblock \emph{Advances in applied mathematics} 6(1): 4--22.

\bibitem[{Lakshminarayanan, Pritzel, and
  Blundell(2017)}]{lakshminarayanan2017simple}
Lakshminarayanan, B.; Pritzel, A.; and Blundell, C. 2017.
\newblock Simple and scalable predictive uncertainty estimation using deep
  ensembles.
\newblock In \emph{Advances in neural information processing systems},
  6402--6413.

\bibitem[{Larochelle et~al.(2007)Larochelle, Erhan, Courville, Bergstra, and
  Bengio}]{larochelle2007empirical}
Larochelle, H.; Erhan, D.; Courville, A.; Bergstra, J.; and Bengio, Y. 2007.
\newblock An empirical evaluation of deep architectures on problems with many
  factors of variation.
\newblock In \emph{Proceedings of the 24th international conference on Machine
  learning}, 473--480.

\bibitem[{Lillicrap et~al.(2015)Lillicrap, Hunt, Pritzel, Heess, Erez, Tassa,
  Silver, and Wierstra}]{lillicrap2015continuous}
Lillicrap, T.~P.; Hunt, J.~J.; Pritzel, A.; Heess, N.; Erez, T.; Tassa, Y.;
  Silver, D.; and Wierstra, D. 2015.
\newblock Continuous control with deep reinforcement learning.
\newblock \emph{arXiv preprint arXiv:1509.02971} .

\bibitem[{Mavrin et~al.(2019)Mavrin, Zhang, Yao, Kong, Wu, and
  Yu}]{mavrin2019distributional}
Mavrin, B.; Zhang, S.; Yao, H.; Kong, L.; Wu, K.; and Yu, Y. 2019.
\newblock Distributional reinforcement learning for efficient exploration.
\newblock \emph{arXiv preprint arXiv:1905.06125} .

\bibitem[{Mnih et~al.(2015)Mnih, Kavukcuoglu, Silver, Rusu, Veness, Bellemare,
  Graves, Riedmiller, Fidjeland, Ostrovski et~al.}]{mnih2015human}
Mnih, V.; Kavukcuoglu, K.; Silver, D.; Rusu, A.~A.; Veness, J.; Bellemare,
  M.~G.; Graves, A.; Riedmiller, M.; Fidjeland, A.~K.; Ostrovski, G.; et~al.
  2015.
\newblock Human-level control through deep reinforcement learning.
\newblock \emph{nature} 518(7540): 529--533.

\bibitem[{Moerland, Broekens, and Jonker(2017)}]{moerland2017efficient}
Moerland, T.~M.; Broekens, J.; and Jonker, C.~M. 2017.
\newblock Efficient exploration with double uncertain value networks.
\newblock \emph{arXiv preprint arXiv:1711.10789} .

\bibitem[{Osband et~al.(2016)Osband, Blundell, Pritzel, and
  Van~Roy}]{osband2016deep}
Osband, I.; Blundell, C.; Pritzel, A.; and Van~Roy, B. 2016.
\newblock Deep exploration via bootstrapped DQN.
\newblock In \emph{Advances in neural information processing systems},
  4026--4034.

\bibitem[{Pathak et~al.(2017)Pathak, Agrawal, Efros, and
  Darrell}]{pathak2017curiosity}
Pathak, D.; Agrawal, P.; Efros, A.~A.; and Darrell, T. 2017.
\newblock Curiosity-driven exploration by self-supervised prediction.
\newblock In \emph{Proceedings of the IEEE Conference on Computer Vision and
  Pattern Recognition Workshops}, 16--17.

\bibitem[{Schulman et~al.(2017)Schulman, Wolski, Dhariwal, Radford, and
  Klimov}]{schulman2017proximal}
Schulman, J.; Wolski, F.; Dhariwal, P.; Radford, A.; and Klimov, O. 2017.
\newblock Proximal policy optimization algorithms.
\newblock \emph{arXiv preprint arXiv:1707.06347} .

\bibitem[{Smith(2001)}]{smith2001disentangling}
Smith, L.~A. 2001.
\newblock Disentangling uncertainty and error: On the predictability of
  nonlinear systems.
\newblock In \emph{Nonlinear dynamics and statistics}, 31--64. Springer.

\bibitem[{Strehl and Littman(2005)}]{strehl2005theoretical}
Strehl, A.~L.; and Littman, M.~L. 2005.
\newblock A theoretical analysis of model-based interval estimation.
\newblock In \emph{Proceedings of the 22nd international conference on Machine
  learning}, 856--863.

\bibitem[{Sutton and Barto(2018)}]{sutton2018reinforcement}
Sutton, R.~S.; and Barto, A.~G. 2018.
\newblock \emph{Reinforcement learning: An introduction}.
\newblock MIT press.

\bibitem[{Tang et~al.(2017)Tang, Houthooft, Foote, Stooke, Chen, Duan,
  Schulman, DeTurck, and Abbeel}]{tang2017exploration}
Tang, H.; Houthooft, R.; Foote, D.; Stooke, A.; Chen, O.~X.; Duan, Y.;
  Schulman, J.; DeTurck, F.; and Abbeel, P. 2017.
\newblock \# exploration: A study of count-based exploration for deep
  reinforcement learning.
\newblock In \emph{Advances in neural information processing systems},
  2753--2762.

\bibitem[{Thrun and Schwartz(1993)}]{thrun1993issues}
Thrun, S.; and Schwartz, A. 1993.
\newblock Issues in using function approximation for reinforcement learning.
\newblock In \emph{Proceedings of the 1993 Connectionist Models Summer School
  Hillsdale, NJ. Lawrence Erlbaum}.

\bibitem[{Todorov, Erez, and Tassa(2012)}]{todorov2012mujoco}
Todorov, E.; Erez, T.; and Tassa, Y. 2012.
\newblock Mujoco: A physics engine for model-based control.
\newblock In \emph{2012 IEEE/RSJ International Conference on Intelligent Robots
  and Systems}, 5026--5033. IEEE.

\bibitem[{Van~Hasselt, Guez, and Silver(2016)}]{van2016deep}
Van~Hasselt, H.; Guez, A.; and Silver, D. 2016.
\newblock Deep reinforcement learning with double q-learning.
\newblock In \emph{Thirtieth AAAI conference on artificial intelligence}.

\bibitem[{Watkins and Dayan(1992)}]{watkins1992q}
Watkins, C.~J.; and Dayan, P. 1992.
\newblock Q-learning.
\newblock \emph{Machine learning} 8(3-4): 279--292.

\end{thebibliography}
\bibliographystyle{aaai}

\clearpage

\appendix
\section{Additional Details for Toy Environment}
\label{Sec:Details for Toy Env}


The episode starts with setting the penguin at the bottom grid and terminates when the penguin reaches the fish grid or the elapsed time steps exceed 200. The agent receives a negative reward (-1) at each time step, which encourages the penguin to finish the task as soon as possible. The fire grids are not accessible, and every time the penguin tries to enter the fire grid, it will receive a large negative reward (-3). When the penguin catches the fish, it will receive a large reward (100). Transitions in all the grids are deterministic except the storm grid at the center, in which the action selected has a 75\% chance of being overridden by a random action. To perform continuous control in this grid world environment, we map the output of algorithms into four discrete intervals representing the four moving directions. For the implementations of TD3 and DDPG, we use the code provided by the authors of TD3 \cite{fujimoto2018addressing}.

\section{Proofs}

\subsection{The Insufficient Exploration of TD3}
\label{Sec:TD3 derivation}
\textbf{Lemma 1} The \textit{minimum} operation in target value computation of TD3 can be represented in the mathematical form: 
\begin{align}
min(Q_{\theta 1}, Q_{\theta 2}) \nonumber = \mu(Q_{\theta 1}, Q_{\theta 2}) - \sigma(Q_{\theta 1}, Q_{\theta 2}) \nonumber 
\end{align}
where $\mu$ and $\sigma$ represent the mean and stand derivation of $Q_{\theta 1}$ and $Q_{\theta 2}$, respectively. \\ \\
\textbf{Proof:} Writing $\mu(Q_{\theta 1}, Q_{\theta 2})$ as $\mu$, and $\sigma(Q_{\theta 1}, Q_{\theta 2})$ as $\sigma$ for short, we have: 
\begin{align}
min(Q_{\theta 1}&, Q_{\theta 2}) = \mu - (\mu - min(Q_{\theta 1}, Q_{\theta 2})) \nonumber \\
  &= \mu - \sqrt{\frac{1}{2}*2(\mu - min(Q_{\theta 1}, Q_{\theta 2}))^2}
\label{EQ:TD3_EXP_0}
\end{align}
As $\mu = \frac{1}{2}(max(Q_{\theta 1}, Q_{\theta 2}) + min(Q_{\theta 1}, Q_{\theta 2}))$, we have 
\begin{align}
&2\mu = max(Q_{\theta 1}, Q_{\theta 2}) + min(Q_{\theta 1}, Q_{\theta 2}) \\
&\mu - min(Q_{\theta 1}, Q_{\theta 2}) = max(Q_{\theta 1}, Q_{\theta 2}) - \mu \\
&(\mu - min(Q_{\theta 1}, Q_{\theta 2}))^2 = (max(Q_{\theta 1}, Q_{\theta 2}) - \mu) ^ 2 
\label{EQ:TD3_EXP_1}
\end{align}
According to Equation~\ref{EQ:TD3_EXP_1}, we replace the term $2(\mu - min(Q_{\theta 1}, Q_{\theta 2}))^2$ in Equation~\ref{EQ:TD3_EXP_0} with $(\mu - min(Q_{\theta 1}, Q_{\theta 2}))^2 + (max(Q_{\theta 1}, Q_{\theta 2}) - \mu) ^ 2$, then obtain:
\begin{align*}
min(Q_{\theta 1}, Q_{\theta 2}) &= \mu - \sqrt{\frac{1}{2}((\mu - min(Q_{\theta 1}, Q_{\theta 2}))^2 +}\\
&\quad \quad \overline{(\mu - max(Q_{\theta 1}, Q_{\theta 2}))^2)} \\
&= \mu - \sigma
\end{align*}
since $\left\{ Q_{\theta 1}, Q_{\theta 2} \right\} = \left\{min(Q_{\theta 1}, Q_{\theta 2}), max(Q_{\theta 1}, Q_{\theta 2}) \right\}$.\\
\textbf{Q.E.D.}

\subsection{Derivation of ADER}
\label{Sec:ADER derivation}
Let $\alpha>0$ be a constant value that determines penalty on intrinsic uncertainty to reduce the overestimation bias, and $\beta>0$ be a  constant value that indicates the bonus on parametric uncertainty for exploration. Writing the ideal target values as $y_{\text{ideal}} = r + \gamma( \mu - \alpha * \sigma_I + \beta * \sigma_P)$, we wish to find a multiplier $\eta(t)$ for the penalty on estimation uncertainty: $y_{\text{ADER}} = r + \gamma (\mu - \eta(t) * \sigma)$, such that for an arbitrary state-action pair (s,a), the target of ADER is close to the ideal setting (Equation~\ref{EQ:Ideal}). By setting $y_{\text{ADER}} = y_{\text{ideal}}$, we have:
\begin{align}
\mu  -\alpha \sigma_I + \beta \sigma_P &= \mu -\eta(t)\sigma_I - \eta(t)\sigma_P \nonumber, 
\end{align}
which gives us
\begin{align}
\eta(t) &= \alpha - \frac{(\alpha + \beta)\sigma_P}{\sigma_I + \sigma_P}.
\end{align}
We then define $x=\frac{\sigma_P}{\sigma_I}$. Recall that $\sigma_P$ decreases gradually as more training samples are collected: $\lim\limits_{t \to \infty} x= 0$. We do first-order Taylor expansion at $x = 0$ on $\eta(t)$:
\begin{align}
\eta(t) &= \alpha - (\alpha + \beta) \frac{x}{x + 1} \nonumber \\
&\approx \alpha - (\alpha + \beta)x.
\end{align}

\section{Implementation Details}
\label{sec:hyper-parameters}
The implementation of ADER is based on the code provided by the TD3 authors (https://github.com/sfujim/TD3). Our modification is simple to replicate as we only upgrade the computation of target values, according to Equation~\ref{EQ:ADER_schema}.
The full hyper-parameters used in our experiment are listed in Table~\ref{table:hyper-parameters}.

\begin{table}[h]
\begin{center}
\begin{small}
\begin{tabular}{| c | c | c |}
\toprule
\bf{Hyper-parameter} & \bf{Value}&\bf{Notes}\\
\midrule
optimizer type & Adam & \makecell{optimizer selected for\\ training neural networks}\\
\hline
actor\_lr & 3e-4 & \makecell{learning rate for \\the actor neural network} \\
\hline
critic\_lr & 3e-4 & \makecell{learning rate for \\ the critic neural network} \\
\hline
batch\_size & 256 &  \makecell{batch size for \\ neural network training}\\
\hline
policy\_freq & 2 & \makecell{frequency of delayed \\ policy updates (TD3)}\\
\hline
noise\_clip & 2 & \makecell{range to clip  \\ target policy noise (TD3)}\\
\hline
$\sigma$ & 0.1 & \makecell{standard deviation of \\Gaussian exploration noise}\\
\hline
$\gamma$ & 0.99 & \makecell{discount factor for \\ reward computation} \\
\hline
$\alpha$ & 2.0 & \makecell{the penalty multiplier \\for intrinsic uncertainty} \\
\hline
$\kappa$ & 5.0 & \makecell{the decay rate of bonus \\for parametric uncertainty} \\
\hline
$K$ & 10000 & \makecell{the number of environment\\ step in a epoch} \\
\hline
$M$ & 100,000 & \makecell{train steps elapsed before \\updating $\eta(t)$ again}\\
\hline
$\tau$ & 0.005 & \makecell{update rate \\for the target networks} \\
\bottomrule
\end{tabular}
\end{small}
\end{center}
\caption{Hyper-parameters used in the experiments.}
\label{table:hyper-parameters}
\end{table}

\end{document}